# The Method for Storing Patterns in Neural Networks
## -Memorization and Recall of QR code[1]  Patterns-

Hiroshi Inazawa

*Center for Education in Information Systems, Kobe Shoin Women's University[2],*
*1-2-1 Shinohara-Obanoyama, Nada Kobe 657-0015, Japan.*

## Abstract

In this paper, we propose a mechanism for storing complex patterns within a neural network and subsequently recalling them. This model is based on our work published in 2018(Inazawa, 2018), which we have refined and extended in this work. With the recent advancements in deep learning and large language model (LLM)-based AI technologies (generative AI), it can be considered that methodologies for the learning are becoming increasingly well-established. In the future, we expect to see further research on memory using models based on Transformers (Vaswani, et. al., 2017, Rae, et. al., 2020), but in this paper we propose a simpler and more powerful model of memory and recall in neural networks. The advantage of storing patterns in a neural network lies in its ability to recall the original pattern even when an incomplete version is presented. The patterns we have produced for use in this study have been QR code (DENSO WAVE, 1994), which has become widely used as an information transmission tool in recent years.

---

[1]  QR Code is a registered trademark of DENSO WAVE Inc.
[2]  This position runs until the end of March 2025.

## 1. Introduction

This paper proposes a mechanism for storing and recalling complex patterns within a neural network. The model is based on our work published in 2018 (Inazawa, 2018) and has been improved and extended in this work. In that paper, although a unidirectional learning model was proposed, from a single memory neuron to a group of neurons that represent patterns, we extend this mechanism to a bidirectional learning model in this paper. Until now, memory in neural networks has mainly been studied using associative memory models (Anderson,1972, Kohonen, 1972, Amari, et. Al., 1988, Yoshizawa, et. al., 1993, Inazawa, et. et., 1988, Hopfield, 1982). In the future, it is expected that research on memory will progress further using models based on Transformers (Vaswani, et. al., 2017, Rae, et. al., 2020), but in this paper we propose a simpler and more powerful model of memory and recall in neural networks. In traditional models of associative memory, memory rate has played an important role. This memory rate has commonly been cited as approximately 15% of the total number of neurons used, those in the output layer (Yoshizawa, et. al., 1993, Inazawa, et. et., 1988, Amit, et. al., 1985, Widow, et. al., 1960 and 1988). The model introduced here allows a single neuron to store and retain one pattern[3], resulting in a memory rate of 100%, if we borrow the term "memory rate." The basic structure of the model consists of a group of memorizing neurons responsible for storing multiple memories, and a single-layer neural network for representing patterns. The memory neurons store memories with a one-to-one correspondence between one memory neuron and one pattern, and when these memorizing neurons are activated, the original pattern is recalled. In this system, the number of memorizing neurons increases with the number of patterns to be stored. Specifically, to store N patterns, a total of N+R neurons are required, where R is the fixed number of neurons in the neural network used to display the patterns. In addition, the patterns used in this study are QR code, which have been widely adopted in recent years as information transmission tools, and QR code patterns of 298 sheets are generated in this study. The content of the QR code consists of basic Japanese phonetic characters (246 characters) combined with the alphabet (52 characters) and encoded into a QR code.

One advantage of the method presented in this paper is that it allows recall of the original pattern even when an incomplete pattern is presented. On the other hand, a drawback is that as the number of patterns to be stored increases, the number of neurons also increases, which in turn results in a larger memory usage. More specifically, using $N + R$ neurons result in a total of $N \cdot R$ synaptic weights. When the number of patterns to be stored becomes extremely large, practical issues regarding memory usage are likely to arise. However, considering the recent advancements in large-capacity and high-performance devices, it can be assumed that a certain level of practicality can still be maintained. In Section 2, the details of the model will be presented, followed by the simulation methods and results in Section 3. Finally, Section 4 will discuss the conclusions and provide a discussion.

---

[3] A grandmother cell has previously been proposed, i.e., where one neuron is responsible for memory.

## 2. Specification of the model

In this section, we provide a more detailed explanation of the model's structure, keeping in mind the simulations will discuss in Section 3. The model structure consists of a neural network consisting of a variable number of groups of neurons (called M-Net) responsible for memorizing patterns and a group of 13456 fixed neurons (called R-Net) responsible for displaying patterns.

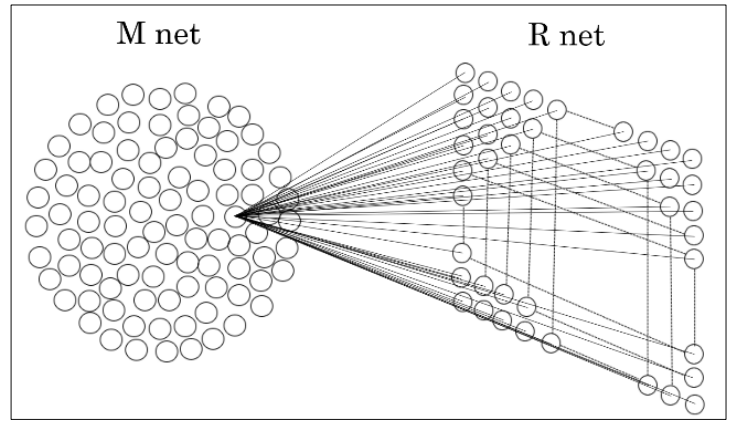

**Figure.1.** A schematic diagram of the proposed model's neuronal group structure. Each neuron in the M-Net is connected to every neuron in the R-Net. With the mechanism, a single QR code pattern displayed by the R-Net is memorized in one neuron of the M-Net.

The number of neurons in R-Net (=13456), is determined by the total number of bits in the QR code pattern, which consists of 116 bits × 116 bits. Since an M-Net neuron memorizes one pattern, the number of memory neurons corresponds to the number of patterns to memorize. M-Net neurons can be added as needed, allowing for flexible scalability. Although theoretically an unlimited number of M-Net neurons can be used, practical constraints such as memory efficiency constrain us to use a finite number of neurons. In Figure 1, we show that each neuron in M-Net is connected to every neuron in R-Net, where one QR code pattern presented to R-Net is memorized in one corresponding M-Net neuron as explained above. Note that there are no connections between the neurons within the M-Net. The learning for M-Net neurons is performed bidirectionally between a single M-Net neuron and all neurons in the R-Net. As the learning method for memorizing, a standard gradient descent method (GDM) is used (Widow, et. al., 1960 and 1988, Reslora, et. al., 1972, Rumelhart, et. al., 1986).

First, we explain how M-Net neurons learn memory. In this memory training, a QR code pattern is first presented to the R-Net, and learning is performed between one M-Net neuron and all neurons in the R-Net using GDM. This is carried out using the input-output relationship between M-Net neurons and R-Net neurons, along with the error function $E^p$. The input-output relationship and the error function are well-known and can be expressed as follows:

$$y_j^p = w_{ji}^p x_i^p \ , \qquad E^p \equiv \frac{1}{2} \sum_{j=0}^{R-1} \left( d_j^p - y_j^p \right)^2 \ , \tag{1}$$

where, $p$ denotes the pattern number, $y_j^p$ is the output of the $j\,th$ R-Net neuron, $x_i^p$ is the input from the $i\,th$ M-Net neuron, and $w_{ji}^p$ is the connection weight from the $i\,th$ M-Net neuron to the $j\,th$ R-Net neuron, and $d_j^p$ is the value of the $j\,th$ of $p\,th$ pattern. $R$ represents the number of neurons in the R-Net. It should be noted that in Equation (1), there is no summation over "$i$." The update of the connection weight $w_{ji}^p$ via Gradient Descent Method (GDM) is performed using the following equation:

$$w_{ji}^p(t+1) = w_{ji}^p(t) + \varDelta w_{ji}^p(t) \,,$$

$$\varDelta w_{ji}^p(t) \equiv -\varepsilon_W \frac{\partial E^p}{\partial w_{ji}^p(t)} = \varepsilon_W \left( d_j^p - y_j^p(t) \right) x_i^p(t), \tag{2}$$

where $t$ represents the number of updates, and $\varepsilon_W$ is the learning rate of M-Net neurons. In learning the connection weight $w_{ji}^p$, the input $x_i^p$ from the M-Net neuron is always set to "1." Also, the initial value of $w_{ji}^p$ at $t = 0$ is set to $0.0$. When $\varepsilon_W$ is set to 1.0, equations (1) and (2) lead to the following relationship between $y_j^p$ and $d_j^p$:

$$y_j^p(t+1) = d_j^p \tag{3}$$

From equation (3), it can be seen that after the learning of $w_{ji}^p$, the output value $y_j^p$ of the R-Net neuron matches the element value $d_j^p$ of the presented pattern. Next, let us explain the learning process of the connection weights $v_{ij}^p$ on the M-Net neuron. The input-output relationship from the R-Net neurons to the $i\ th$'s M-Net neuron can be defined as follows:

$$q_i^p = \sum_{j=0}^{R-1} v_{ij}^p y_j^p \implies x_i^p = \begin{cases} 1.0 & \text{for } max(q_i^p) \\ 0.0 & \text{for } others \end{cases}, i = p = 0 \sim N_P - 1, \tag{4}$$

where $v_{ij}^p$ is the connection weight from the $j\ th$'s R-Net neuron to the $th$'s M-Net neuron, $q_i^p$ represents the output value of the $i\ th$'s M-Net neuron, and $N_P (= 298)$ is the total number of patterns to be presented. The output value $x_i^p$ of the M-Net neuron that represents the maximum value of $q_i^p$ is set to 1.0, and the outputs of all other neurons are set to "0.0." Note that the error function of the M-Net neuron is defined as follows:

$$e^p \equiv \frac{1}{2} \sum_{i=0}^{N_p-1} \left(\theta - q_i^p\right)^2, \tag{5}$$

where $\theta$ serves as the teacher signal for the M-Net neuron. Note that in the error function defined in equation (5), the output values $q_i^p$ of the M-Net neurons defined in equation (4) are used instead of $x_i^p$. Then, the update of the connection weights $v_{ij}^p$ using GDM is done as follows, similar to equation (2):

$$v_{ij}^p(t'+1) = v_{ij}^p(t') + \Delta v_{ij}^p(t'),$$

$$\Delta v_{ij}^p(t') = -\varepsilon_V \frac{\partial e^p}{\partial v_{ij}^p} = \varepsilon_V \left(\theta - q_i^p(t')\right) y_j^p(t'), \tag{6}$$

where $t'$ represents the number of updates, and $\varepsilon_V$ represents the learning rate of R-Net neurons.

## 3. Simulation and results

This section discusses the simulation method based on the model described in Section 2, as well as the results obtained from it. The QR code pattern used in the simulation consists of 298 patterns produced specifically for this purpose. These patterns are QR codes for basic Japanese phonograms (246 characters) and the English alphabet (52 characters). A sample is shown in Figure 2.

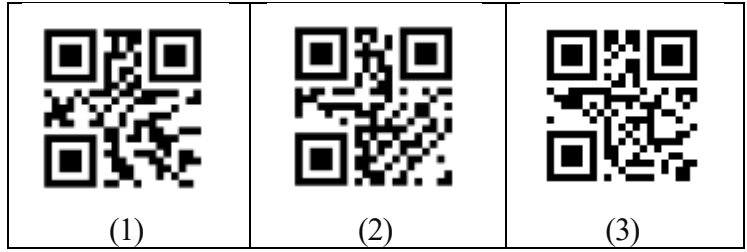

**Figure 2**: An example of a QR code generated for use in the simulation. Each pattern is a 116x116 pixel image. The sample QR code in this figure represents the following string: (1) "あ", (2) "あ", and (3) "A" repeated multiple times, where (1) and (2) are the Japanese katakana and hiragana characters, respectively.

When inputting the QR pattern shown in Figure 2 into R-Net, data in which the values of each point that makes up the pattern have been converted into digital values is used.

First, the memorization learning algorithm for $w_{ji}^p$ in equations (1) and (2) is provided below.

1. Initialization of input and output values in equation (1): $x_i^p = 1.0$, $y_j^p = 0.0$

2. Initialization of connection weights in R-Net: $w_{ji}^p = 0.0$

3. Sequentially retrieve the QR code patterns (digital value patterns) starting from the 0th pattern.

4. Learning of connection weights according to equation (2): $w_{ji}^p(t+1) = w_{ji}^p(t) + \Delta w_{ji}^p(t)$

5. Repeat the above process (steps 3–4) for each pattern: $p = 0 \sim 297$

6. After all patterns have been learned, proceed to process step "7."

7. Once all patterns have been learned, write the learned $w_{ji}^p$ values to a file.

Next, the memorization learning algorithm for $v_{ij}^p$ in equations (4) and (6) is presented.

1. Initialization of the connection weights, input values, and output values in M-Net: $v_{ij}^p = 0.0$, $x_i^p = 1.0$, $y_j^p = 1.0$

2. Load the pre-learned connection weights $w_{ji}^p$.

3. Calculate $y_j^p$ using equation (1)

4. Learning of connection weights according to equation (6): $v_{ij}^p(t'+1) = v_{ij}^p(t') + \Delta v_{ij}^p(t')$: $q_i^p = 0.0$

5. Repeat the above process (steps 3–5) for each pattern: $p = 0 \sim 297$

6. After all patterns have been learned, proceed to process step "7."

7. Once all patterns have been learned, write the learned $v_{ij}^p$ values to a file.

Meanwhile, in the recall process, the digital values of the memorized pattern calculated using equation (1) is presented to the R-Net, and the output value $q_i^p$ of each M-Net neuron (i = $0 \sim 298$) is calculated using equation (4). The M-Net neuron number that outputs the maximum value among these output values $q_i^p$ indicates the memorized pattern. The algorithm of the recall process can be summarized as follows:

1. Load the pre-trained connection weights $w_{ji}^p$.

2. Load the pre-trained connection weights $v_{ij}^p$.

3. Select a trained arbitrarily pattern.

4. Calculate $y_j^p$ using equation (1): where $x_i^p = 1.0$.

5. For all M-Net neurons, calculate $q_i^p$ using Equation (4).

6. Perform normalization $q_i^p / R (\equiv \text{Normalized\_} q_i^p)$.

※ Note that from now on, when we refer to $q_i^p$, we mean Normalized_$q_i^p$.

7.  Determine the maximum value of $q_i^p$ among i=0~297.

8.  Identify the pattern number corresponding to the maximum value obtained in Step 7, and calculate $y_j^p$ using equation (1). Display this pattern on the R-Net.

9.  Repeat the above process from Step 3 to Step 7 while changing the selected pattern.

Now, Figure 3 shows the firing patterns of M-Net neurons when we have performed the above recall process. The presented pattern (memory pattern) corresponds to the 145th QR code, consisting of eight repetitions of the character "ぎ".

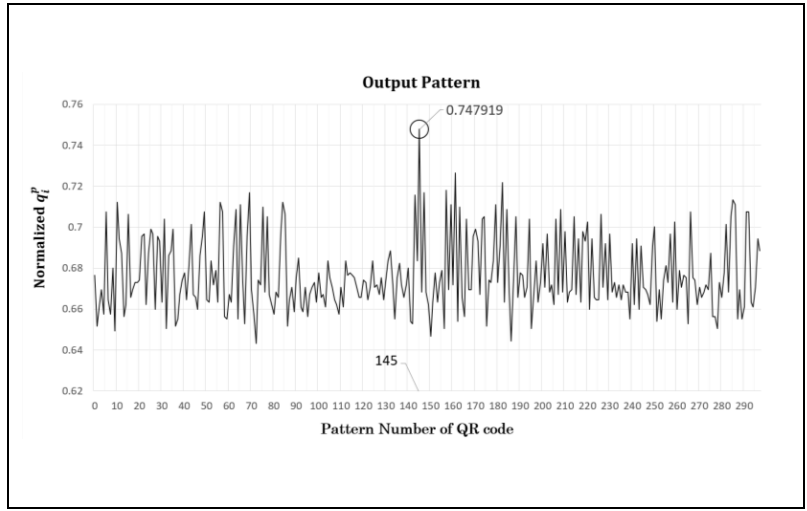

**Figure 3**: Outputs of M-Net neurons. This shows the output $q_i^p$ of the 0th to 297th M-Net neurons. The presented pattern is the 145th pattern. Its output value is 0.747919. The inserted image surrounded by a gray square is the presented QR code pattern.

Then, based on the recall algorithm described above, all stored patterns (neuron numbers) were compared with the presented patterns to examine the consistency between the stored neuron numbers and the corresponding presented pattern numbers. As a result, the stored neuron numbers and their corresponding presented pattern numbers have been found to match perfectly for all 289 patterns. This has been confirmed that all pattern numbers and their corresponding patterns have been accurately recalled without exception. Figure 4 shows the normalized maximum output values, Max $q_i^p$, for all 289 patterns.

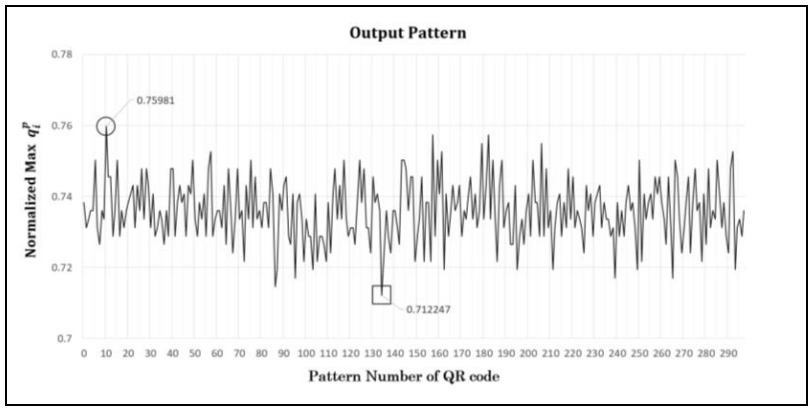

**Figure 4**: Outputs of M-Net neurons. This shows the output of the 0th to 297th M-Net neurons. The vertical axis shows (Normalized) Max $q_i^p$. Each peak value is the (Normalized) Max $q_i^p$ every presented pattern. The maximum value among these was 0.75981, while the minimum value was 0.712247.

Next, we explore whether we can identify the original pattern when only part of it is presented. To perform this, we present R-Net with partially missing patterns, e.g., only the bottom 25% of the right side of the 145th pattern. The results are shown in Figure 5.

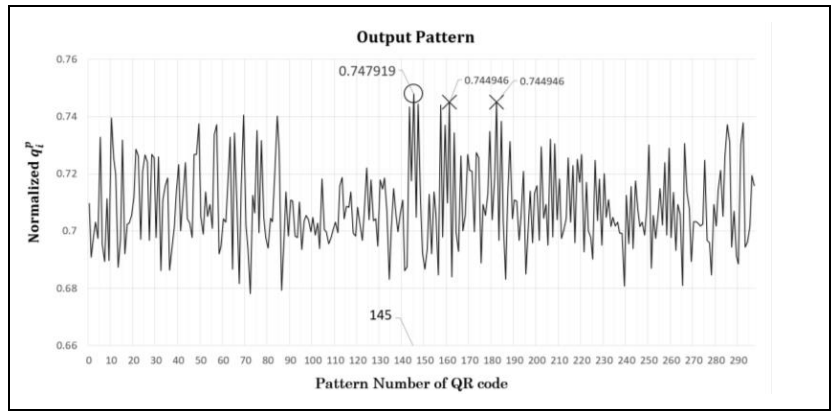

**Figure 5**: Output of M-Net neurons. This shows the output of the 0th to 297th M-Net neurons. The vertical axis shows $q_i^p$. The presented pattern is 25% of the lower right part of the 145th pattern. The maximum output value has been 0.747919, which is the 145th neuron. This result is identical to the output value obtained when the entire pattern is presented (see Figure 3). The second highest values are 0.744946 at 161st and 182nd. The inset image in the grey box shows the 145 th pattern presented at 25%.

In this case, the maximum output value was the same as when 100% of the 145th presented pattern was shown, with $q_i^p = 0.747919$. Since the presented pattern is only 25% of the original, many similar patterns are generated, resulting in a larger number of patterns showing values near the maximum value. The second-largest output is shown by the 161st and 182nd patterns, with $q_i^p = 0.744946$. These results show that even when a pattern is presented partially, recall is still possible. Furthermore, when the presented pattern is reduced to 20.7% or less of the original, the output value corresponding to a completely different pattern has been become the maximum, indicating that recall failed in this case. However, it is important to note that this result of 20.7% is not significant in itself, as it is affected by factors such as the density and complexity of the patterns. The important point is that it is possible to successfully recall an incomplete image.

Finally, let's look at the output values of M-Net neurons when presented with a completely unknown pattern that has never been memorized before. The unmemorized pattern has replaced by replacing the 145th pattern, "ぎ," with a QR code pattern representing "あいうえお." The output pattern in this case is shown in Figure 6. The output values have been all smaller than 0.7. The output value of the 145th pattern presented was also $q_i^p = 0.674197$, which is significantly smaller than the other output values. Naturally, even in such cases, there exists an M-Net neuron with the maximum output value ($q_i^p = 0.697979$), however, this is a pattern entirely different from the presented one. Therefore, it is reasonable to set a certain threshold as a criterion for determining whether memory recall has completely failed. As shown in Figure 4, when the memorized patterns match the M-Net neuron numbers, their output values are all above 0.7. Therefore, a threshold of 0.70 is considered appropriate for this model.

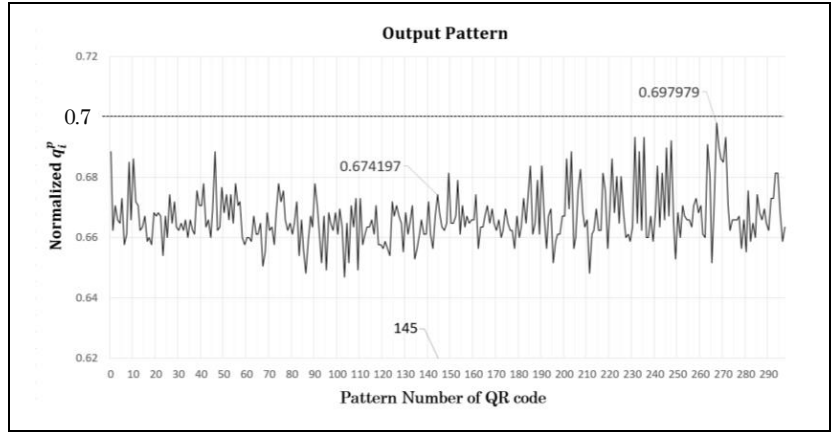

**Figure 6**: M-Net neuron output. This shows the output $q_i^p$ of M-Net neurons 0th to 297th. The vertical axis shows "normalized $q_i^p$". The presented pattern is the 145th pattern replaced by an unlearned pattern. The maximum value is the 267th pattern, and the output value at this time is 0.697979. The horizontal dotted line at $q_i^p$=0.7 shows the threshold for whether the presented pattern is the memorized pattern. We can see that the output does not exceed the threshold for all M-Net neurons.

### 4. Conclusions and discussion

Building on the previously introduced study (Inazawa, 2018), this paper has proposed a mechanism for storing and recalling QR code patterns within a neural network. In this simulation performed in this paper, 298 M-Net neurons were assigned to store one pattern each (298 presented patterns). When all patterns were presented to the R-Net in order, the output value of the M-Net neurons that remembered the presented patterns has been "$q_i^p \gtrsim 0.71$". Among these values, the maximum value of $q_i^p = 0.75981$, and the minimum value of $q_i^p = 0.712247$. Furthermore, when presented with memory patterns that are missing about 75% of the data, M-Net neurons that stored the corresponding patterns still produced the largest output values. This demonstrates that it is possible to completely recall the entire pattern from the missing data. Moreover, when the presented pattern is 20.7% of the original (with 79.3% missing), the output value of a completely different pattern became the highest, indicating a failure in recall. The value (20.7%) of this defect level is also affected by the complexity of the pattern and the density of the dots that make up the pattern, so this value is specific to the pattern used in this simulation. In addition, we also have investigated the case where a pattern that is not stored in the system at all was presented. In this case, no outstanding output values of the M-Net neurons has been observed, and the pattern with a number completely different from the presented pattern has showed the highest output value. Furthermore, the output values for all patterns has been below 0.7 as shown in Figure.6.

One advantage of using neural networks for memory is that even if we are presented with an incomplete input, we can still recall the original pattern. As shown in Section 3, we have found that the model has the ability to reconstruct the overall pattern even when approximately 75% of the presented (memorized) pattern is missing. On the other hand, the major drawback is that as the number of patterns to be stored increases, the number of neurons required also increases, and thus the number of binding constants to be memorized also increases. In this study, the total size of the prepared QR code pattern was 4.65 MB on disk, while the required storage space for the connection weights $w_{ji}^p$ and $v_{ij}^p$ was 7.65 MB each (total 15.3 MB). The recent advancements in deep learning and large language model (LLM)-based artificial intelligence, particularly generative AI, have been remarkable. In terms of system learning, the methodologies involved can be considered increasingly well-established. On the other hand, when it comes to remembering the processes that AI has performed, complete reproducibility may not yet be guaranteed. Of course, it can be assumed that generative AI will produce similar results if tasked with performing the same process. However, the results may not be entirely identical. In the future, models like the Transformer (Vaswani, et. al., 2017, Rae, et. al., 2020) may make memory research increasingly important. On the other hand, from the 1970s to the 2000s, research into associative memory models was active alongside learning models, but in recent years research into the latter may have been less frequent than research into learning. This paper is a model of memory and recall using neural networks that uses a traditional method rather than the latest LLM. We hope that the system proposed here will contribute to the field of memory (data storage) and recall.